\documentclass[11pt]{article}

\usepackage[utf8]{inputenc}
\usepackage[T1]{fontenc}
\usepackage{microtype} 
\usepackage{times} 
\usepackage{amsmath, amssymb, amsfonts}
\usepackage{graphicx}
\usepackage{lineno}
\usepackage[margin=1in]{geometry}
\usepackage{hyperref}
\usepackage{url}
\usepackage{booktabs} 
\usepackage{caption}
\usepackage{authblk}
\usepackage{enumitem} 

\hypersetup{
    colorlinks=true,
    linkcolor=blue,
    filecolor=magenta,
    urlcolor=cyan,
    citecolor=blue, 
    pdftitle={Graph RAG for Legal Norms: A Hierarchical and Temporal Approach},
    pdfpagemode=FullScreen,
}

\title{An Ontology-Driven Graph RAG for Legal Norms: \\ A Structural, Temporal, and Deterministic Approach}

\author[1]{Hudson de Martim}
\affil[1]{Federal Senate of Brazil \\ \texttt{hudsonm@senado.leg.br}}
\date{} 

\begin{document}

\maketitle

\begin{abstract}
Retrieval-Augmented Generation (RAG) systems in the legal domain face a critical challenge: standard, flat-text retrieval is blind to the hierarchical, diachronic, and causal structure of law, leading to anachronistic and unreliable answers. This paper introduces the \textbf{Structure-Aware Temporal Graph RAG (SAT-Graph RAG)}, an ontology-driven framework designed to overcome these limitations by explicitly modeling the \textbf{formal structure} and diachronic nature of legal norms. We ground our knowledge graph in a formal, LRMoo-inspired model that distinguishes abstract legal Works from their versioned Expressions. We model temporal states as efficient aggregations that reuse the versioned expressions (CTVs) of unchanged components, and we reify legislative events as first-class \texttt{Action} nodes to make causality explicit and queryable. This structured backbone enables a unified, planner-guided query strategy that applies explicit policies to deterministically resolve complex requests for (i) point-in-time retrieval, (ii) hierarchical impact analysis, and (iii) auditable provenance reconstruction. Through a case study on the Brazilian Constitution, we demonstrate how this approach provides a verifiable, temporally-correct substrate for LLMs, enabling higher-order analytical capabilities while drastically reducing the risk of factual errors. The result is a practical framework for building more trustworthy and explainable legal AI systems.
\end{abstract}

\noindent\textbf{Keywords:} Graph RAG; Legal Knowledge Graphs; Temporal Modeling; Provenance Reconstruction; Computational Law; Legal Ontology; Deterministic Retrieval; LRMoo

\section{Introduction}
Artificial Intelligence (AI) systems, particularly those based on Retrieval-Augmented Generation (RAG) \cite{lewis2020rag}, offer immense potential for navigating the complexity of the legal domain. However, legal corpora present unique structural and temporal challenges that naive RAG approaches fail to address. Legal norms are not flat documents; they are characterized by a formal hierarchy (titles, chapters, articles), a dense web of cross-references, and, most critically, a continuous diachronic evolution through amendments, repeals, and consolidations.

This temporal dynamism is a fundamental stumbling block for standard AI systems. A system that is "temporally-naïve" cannot deterministically retrieve the version of a law that was valid on a specific historical date, leading to anachronistic and factually incorrect answers. This is unacceptable in a high-stakes domain where precision and auditability are paramount. To build trustworthy legal AI, we need a retrieval substrate that explicitly models the law's structure and evolution. As argued in \cite{demartim2025temporal}, a formal, component-level versioning model is a necessary prerequisite.

To address this gap, this paper introduces an ontology-driven Graph RAG framework specifically designed for the structural and temporal complexities of statutory law. We depart from standard Graph RAG \cite{edge2024graphrag} by leveraging the curated, intrinsic hierarchy of legal norms as our primary community structure, rather than relying on algorithmic community detection. Our core contribution is a knowledge graph grounded in a formal, LRMoo-inspired ontology \cite{lrmoo2024} that models:
\begin{itemize}
    \item \textbf{A formal, multi-layered representation}, building upon the model in~\cite{demartim2025temporal}, that separates abstract legal \texttt{Works} from their versioned, time-stamped \texttt{Expressions}.
    
    \item \textbf{An efficient aggregation model} for propagating changes hierarchically, which avoids data redundancy by reusing unchanged versioned components.
    
    \item \textbf{The reification of legislative events as first-class, retrievable entities}, making causality a directly queryable aspect of the law's history.
    
    \item \textbf{A multi-aspect retrieval approach} that textualizes structured metadata and events, making the context of a norm as searchable as its content.
\end{itemize}

This structured approach enables deterministic query patterns for complex temporal and analyses that are infeasible for standard RAG systems. To better articulate these contributions, we find it useful to conceptually distinguish two core facets of our framework. We refer to the use of the intrinsic document hierarchy—its titles, chapters, and articles—as the graph's backbone as the \textbf{Structure-Aware Graph RAG} perspective. This contrasts with standard Graph RAG, which derives a semantic hierarchy from content rather than adopting the document's formal schema. Complementing our approach, we refer to the modeling of the diachronic evolution of these structural components as the \textbf{Temporal Graph RAG} perspective. It is the seamless integration of these two perspectives, grounded in a formal ontology, that forms the basis of our unified \textbf{Structure-Aware Temporal Graph RAG (SAT-Graph RAG)} framework.

This paper is structured as follows. Section~\ref{sec:related_work} reviews the literature on legal knowledge graphs, RAG, and temporal modeling, highlighting the gaps our work addresses. Section~\ref{sec:framework} details our proposed ontology-driven framework, from the formal model to the graph construction process and the key mechanisms for handling temporality, causality, and structure-aware retrieval. Section~\ref{sec:case_study} presents a case study on the Brazilian Constitution, demonstrating through qualitative evaluation how our framework handles complex query patterns for point-in-time retrieval, hierarchical impact analysis, and causal-provenance reconstruction. Finally, Section~\ref{sec:discussion} discusses the broader implications, limitations, and practical considerations of the approach, and Section~\ref{sec:conclusion} concludes the paper and outlines future research directions.

\section{Related Work}
\label{sec:related_work}

Research on Retrieval-Augmented Generation (RAG) and its graph-based extensions has advanced rapidly. The original RAG formulation~\cite{lewis2020rag} showed that coupling parametric language models with an external retriever improves performance on knowledge-intensive tasks and provides a principled route to provenance and updateability. Recent work has generalized RAG into graph-centric pipelines (``Graph RAG'') that construct an intermediate knowledge graph to support global, corpus-level sensemaking; these methods are a line of research directly relevant to this paper's proposal~\cite{edge2024graphrag}.

In the legal domain, recent studies have adapted RAG and graph-based solutions for specific tasks. Dedicated benchmarks such as LegalBench-RAG show that the retrieval stage remains a critical bottleneck in RAG pipelines applied to legal texts~\cite{pipitone2024legalbenchrag}. Furthermore, works that enrich retrieval with structural information from legislation (e.g., using graphs of articles and links) demonstrate significant gains in tasks like statutory article retrieval and legal question answering~\cite{louis2023finding, hei2024heterogeneous, ho2025fairuse}.

In parallel, the literature on Knowledge Graphs (KGs) has increasingly focused on the temporal dimension (Temporal Knowledge Graphs—TKGs). Recent surveys and reviews specifically address temporal modeling in KGs and its application to temporal reasoning and QA, highlighting both robust representation techniques and gaps in evaluation and datasets~\cite{cai2024survey}. Notably, reviews on temporal KG modeling conclude that while robust methods for representation exist, there is a scarcity of work that explicitly integrates the hierarchical and cross-referential structure typical of normative texts with secure retrieval mechanisms for provenance and historical versions~\cite{hooshafza2022temporal}.

In the legislative context, well-established standards and modeling initiatives aim to capture legislative changes over time, with Akoma-Ntoso~\cite{palmirani2012legislativexml} being a prominent standard. These works show that representing legislation is not merely about text: it includes hierarchy (parts, chapters, articles), cross-references, amendments, and conditional versions, which demands a KG model with explicit temporal and hierarchical support~\cite{stellato2023legalhtml}.

Despite these advances, we identify three key gaps that motivate our work: (i) most RAG systems applied to the legal domain treat documents as flat textual chunks and rarely integrate the normative hierarchical structure explicitly; (ii) work on TKGs has focused on representation and link prediction, but few address the combination—at scale and with provenance—of normative evolution (article/law versions) with RAG mechanisms; and (iii) there is a lack of benchmarks and datasets that combine legal retrieval tasks with the temporal and hierarchical requirements typical of legislative systems~\cite{pipitone2024legalbenchrag, zhu2025temporal, cai2024survey}.

Our work introduces \emph{Graph RAG for Legal Norms}, an architecture that addresses these gaps by: (a) grounding the graph in a formal, LRMoo-inspired ontology that explicitly represents the hierarchy and versioning of legal texts; (b) enabling deterministic, temporally-aware retrieval with auditable provenance (e.g., ``what was the applicable text on 2018-05-01?''); and (c) supporting both local (article-level) and global (corpus-level) analysis through structure-aware aggregation. This proposal combines principles from Graph RAG with temporal KG modeling and normative engineering standards. In doing so, we aim to directly address gaps (i) and (ii), and lay the groundwork for addressing gap (iii) by creating a set of evaluation tasks that test temporal retrieval and provenance accuracy.

Finally, our framework aligns with a growing body of research proposing hybrid pipelines (KG + RAG + vectors) in the legal domain. This line of work validates the hypothesis that fusing structured and textual representations reduces hallucinations and improves explainability. Our approach can be seen as a sophisticated instantiation of this principle, but one that is specifically designed to address the deep granularity and temporality requirements that legislation imposes, which current hybrid models do not yet fully address~\cite{barron2025bridging}.

\paragraph{Summary} In sum, while a solid foundation of RAG and TKG techniques exists, along with initial applications to the legal domain, a systematic integration of (1) a representation of the norm's \textbf{formal structure} and its versioning, (2) temporally-aware retrieval with provenance, and (3) scalable Graph-RAG pipelines for both local and global questions is still lacking—gaps that the approach proposed in this paper aims to fill.

\section{The Proposed Framework}
\label{sec:framework}

Our framework models the formal-structural, temporal, and linguistic complexities of legal norms through an ontology-driven graph structure. Its design is motivated by a logical progression that addresses the shortcomings of existing Graph RAG approaches when applied to structured, diachronic domains like legal norms.

\subsection{Design Rationale: From Entity-Centric to Structural-Temporal RAG}
\label{sec:design_rationale}

Standard Graph RAG approaches primarily focus on an \textbf{entity-centric model}~\cite{edge2024graphrag}. In this model, the graph is constructed by extracting named entities (e.g., people, organizations) and key concepts from the text, which become nodes, while their co-occurrences or inferred relationships become edges. While powerful for discovering thematic connections within content, this approach is fundamentally blind to the formal, hierarchical architecture of the document itself. For legal texts, where the position of a provision within a chapter or title defines its scope and meaning, this structural blindness is a critical limitation.

Beyond its structural blindness, the standard entity-centric model faces a more fundamental challenge in the statutory domain. Named Entity Recognition (NER) systems are optimized to extract proper nouns (e.g., persons, organizations, locations), which are notably sparse in the abstract and impersonal text of legal norms. The true semantic fabric of law is woven from \textbf{legal concepts} (e.g., "administrative act," "due process"), which are not captured by standard NER models. A direct application of an entity-extraction pipeline would therefore yield a sparse and semantically impoverished knowledge graph. This further justifies our decision to pivot from a content-driven, bottom-up approach to a top-down model that leverages the document's formal structure—an ever-present and high-value feature—as the graph's primary backbone.

Then, the first key design decision of our framework is to adopt a \textbf{structural perspective}. Instead of content-derived entities, we model the intrinsic structure of the legal norm as the graph's backbone. Hierarchical components of the document—such as Titles, Chapters, Articles, and Items—are instantiated as nodes, and their parent-child relationships form the primary edges. This shift allows queries to target not just the text's content, but the architecture of the knowledge itself.

However, a purely structural graph would still be temporally-naïve, representing only a single state of the law and failing to capture its diachronic nature. The final and most critical layer of our design is therefore the introduction of the \textbf{temporal perspective}. We elevate this structural model by ensuring that each version of a structural component over time becomes a distinct, addressable node in the graph. As detailed below, this is achieved by distinguishing abstract components from their time-stamped Temporal Versions (CTVs). It is the synthesis of these structural and temporal dimensions that forms our unified \textbf{Structure-Aware Temporal Graph RAG (SAT-Graph RAG)} model, enabling a retrieval mechanism with the precision and determinism required for the legal domain.

\subsection{Ontological Foundation and Graph Construction}
\label{sec:ontology_construction}

At the core of our framework lies a formal model grounded in the IFLA LRMoo ontology \cite{lrmoo2024}, an object-oriented representation of the IFLA Library Reference Model (LRM). Our choice of LRMoo over other standards like Akoma Ntoso~\cite{palmirani2012legislativexml} is deliberate. While Akoma Ntoso provides a powerful XML schema for encoding the structure of legal documents, it represents the conceptual layers of the FRBR/LRM model (Work, Expression, Manifestation) primarily as metadata identifiers within the document's `<meta>` block. 

Our framework, in contrast, elevates these concepts from metadata tags to first-class ontological entities. We therefore ground our graph structure in the formal model established in~\cite{demartim2025temporal}. This model's direct, structural representation of the Work/Expression distinction is fundamental to our granular temporal modeling approach, as it allows us to build explicit, traversable chains of versions within the graph itself, rather than relying solely on the interpretation of metadata. Following this model, we define four primary entity types that are instantiated as nodes:

\begin{enumerate}
    \item \textbf{Norm (as a Work):} Represents a legal norm as an abstract intellectual creation (e.g., the "1988 Brazilian Federal Constitution"). This corresponds to the \texttt{F1 Work} concept in LRMoo, capturing the norm's identity independent of any specific wording or amendment.

    \item \textbf{Component (as a Component Work):} Represents a hierarchical element within a norm (e.g., a title, chapter, or article) as a distinct abstract concept. Each component is an identifiable part of the main Work that maintains its conceptual identity even as its text evolves.

    \item \textbf{Temporal Version (TV / CTV):} Represents a language-agnostic "temporal snapshot" of a \textit{Norm} or \textit{Component} at a specific point in time. It captures the semantic content and logical structure, corresponding to the LRMoo \texttt{F2 Expression} concept. A Component Temporal Version (CTV) is the TV of a specific Component.

    \item \textbf{Language Version (LV / CLV):} Represents the concrete textual realization of a specific \textit{Temporal Version} in a particular language (e.g., the Portuguese text of the 1988-10-05 version of an article). A Component Language Version (CLV) is the LV of a specific CTV. Each is also an \texttt{F2 Expression} derived from a single TV.
\end{enumerate}

This multi-layered structure allows us to precisely model the "what" (the abstract Norm/Component), the "when" (the Temporal Version), and the "how" (the Language Version with its specific text). For simplicity in the monolingual examples that follow, we will often refer to a single \emph{Version} entity, which pragmatically combines a \emph{Temporal Version} and its corresponding \emph{Language Version}.

\subsection{Graph Construction: From Text to a Structured Knowledge Graph}
\label{sec:graph_construction}

\begin{figure}[htbp]
    \centering
    \includegraphics[width=1.0\textwidth]{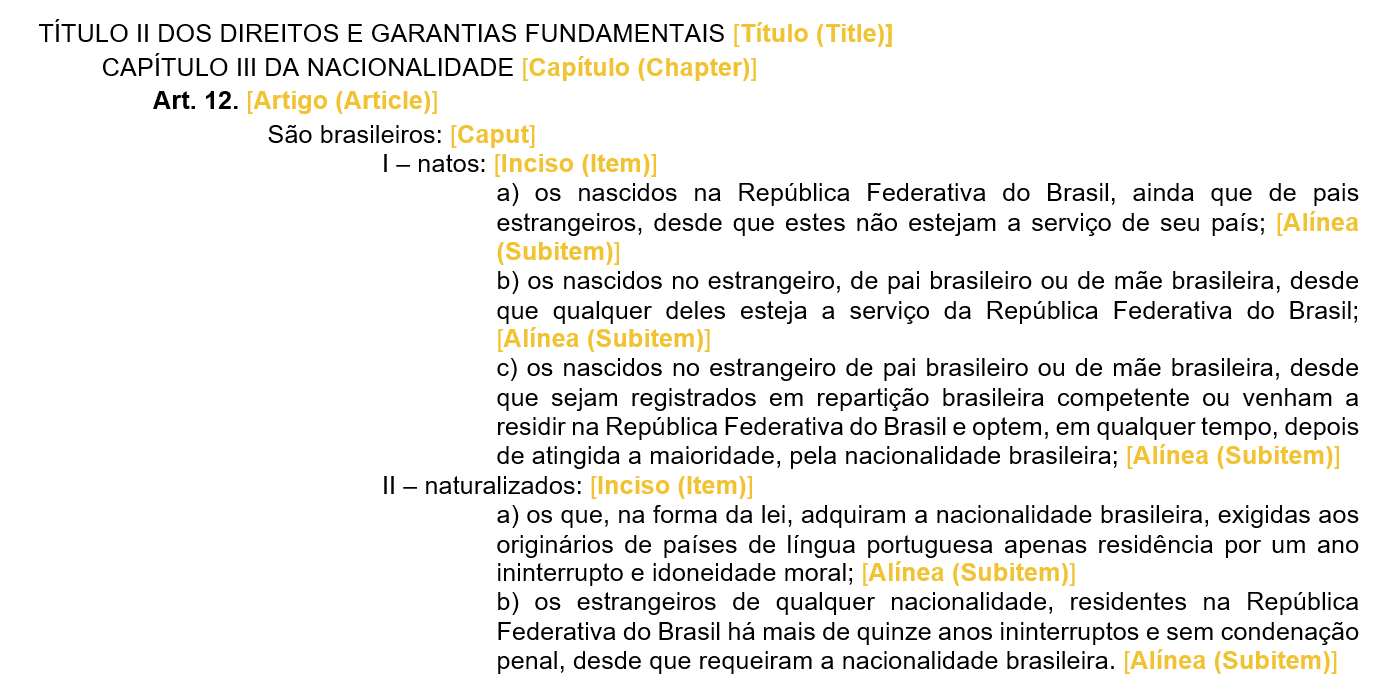} 
    \caption{Example of articulated text for Art. 12 of the Federal Constitution of Brazil (1988) with annotations indicating the types of hierarchical provisions/components.}
    \label{fig:example_art12_n}
\end{figure}

Unlike traditional RAG pipelines that begin with naive text chunking, our process starts with a structure-aware \textbf{semantic segmentation}. The goal is to parse the raw legal text into segments that directly correspond to the norm's intrinsic hierarchical elements (e.g., titles, chapters, articles, paragraphs), as illustrated in Figure~\ref{fig:example_art12_n}. This can be achieved using a specialized parser or a fine-tuned LLM, ideally followed by human review to ensure accuracy.

\begin{figure}[htbp]
    \centering
    \includegraphics[width=0.8\textwidth]{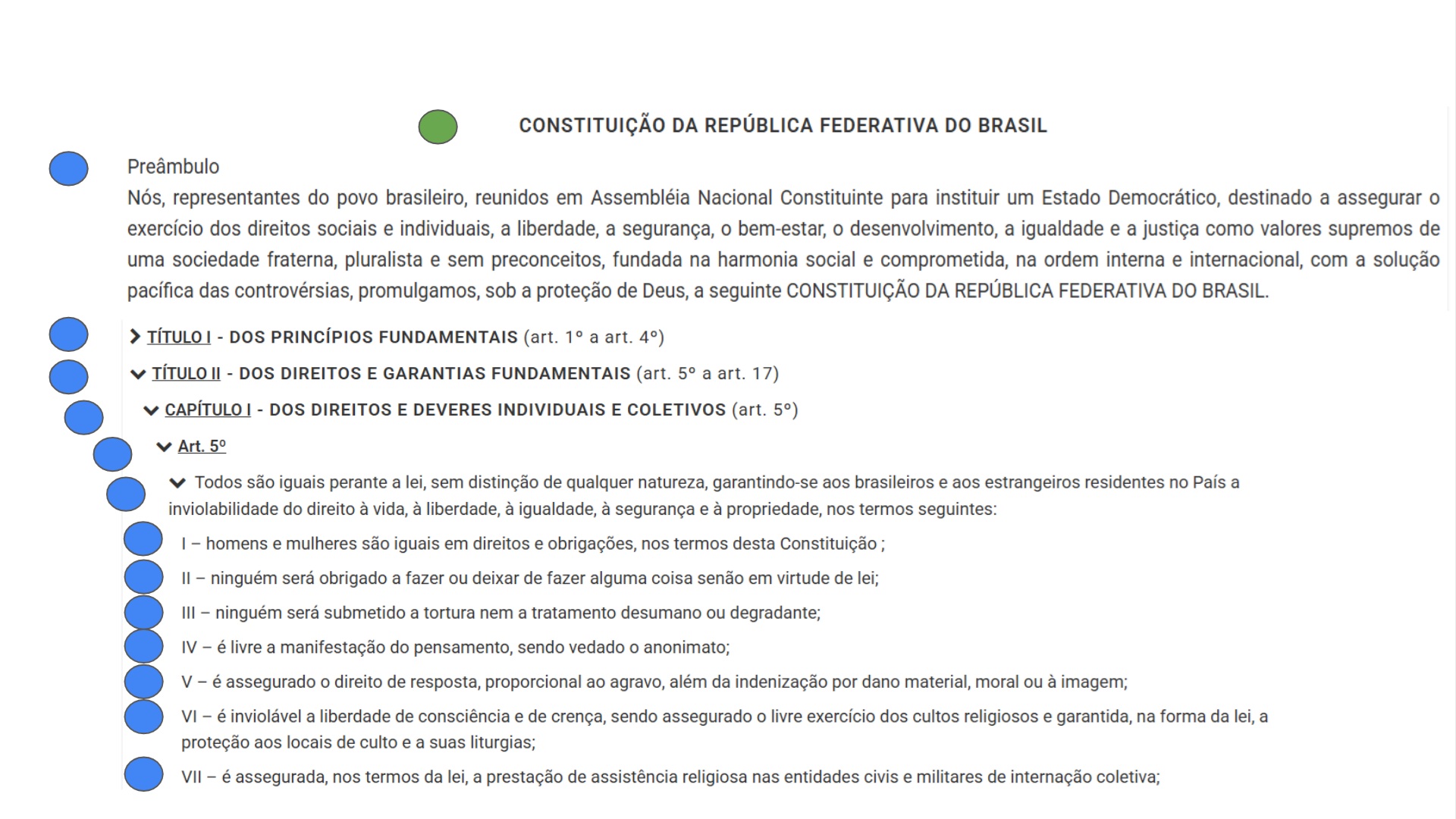} 
    \caption{Illustration of hierarchical semantic segmentation, and typification of the structural entities/nodes,  applied to a passage of the Federal Constitution of Brazil (1988), representing the \emph{Norm} (\textbf{Work}) in green, \emph{Components} (\textbf{Component Works}) in blue.}
    \label{fig:segmentation_example_n}
\end{figure}

This initial step simultaneously identifies and extracts the abstract structural entities. For each legal norm, we create a single \textbf{Norm (Work)} node and multiple \textbf{Component (Work)} nodes, forming the hierarchical backbone of the graph (Figure~\ref{fig:segmentation_example_n}).

The textual content is then linked to this abstract structure. A direct link, however, from a concrete text chunk to a timeless \textbf{Component (Work)} node would be conceptually flawed. Such a link would fail to capture the law's diachronic nature, where the wording of a component evolves with each amendment and can be expressed in multiple languages. The abstract \emph{Component} is permanent, but its textual manifestation is not.

Our model resolves this fundamental challenge by using the versioning layers as the necessary bridge between the abstract and the concrete. For each \emph{Component (Work)}, and for each moment in time its text was enacted or amended, we instantiate its specific realizations by creating:
\begin{enumerate}
    \item A \textbf{Temporal Version (CTV)} node, representing its semantic content on that specific date.
    \item A corresponding \textbf{Language Version (CLV)} node, representing the concrete wording in a specific language.
\end{enumerate}

\begin{figure}[htbp]
    \centering
    \includegraphics[width=0.7\textwidth]{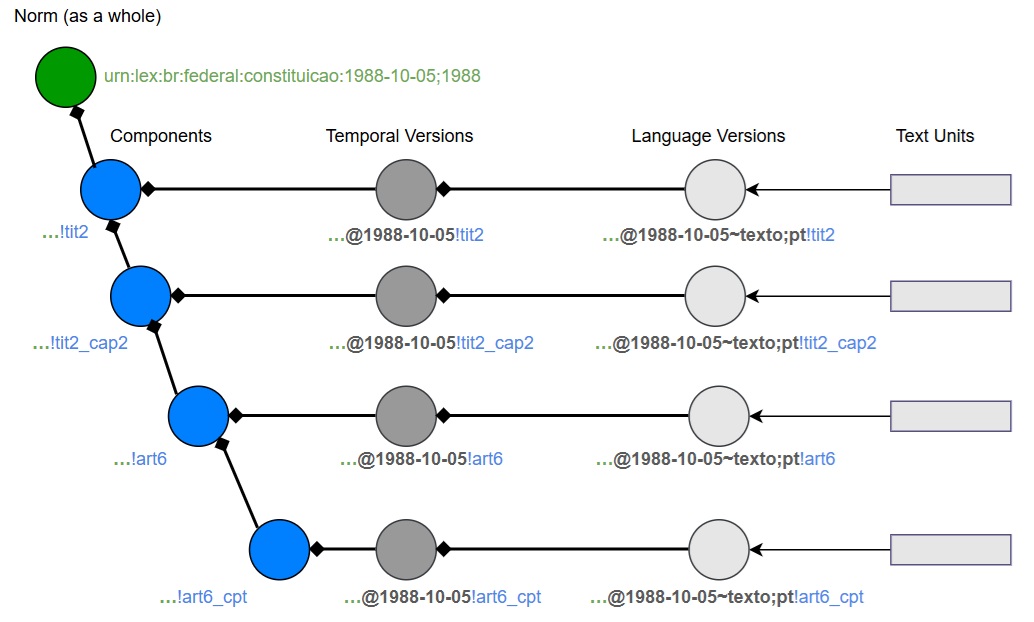} 
    \caption{Representation of the multi-layered relationship in the graph: a \emph{Norm} (Brazilian Constitution of 1988-10-05 (1988) has \emph{Components} in a hierarchy (Title II, Chapter II, Article 6 and its caput), which have date-stamped \emph{Temporal Versions} (CTVs) (all from the original version), which in turn have language-specific (in Portuguese) \emph{Language Versions} (CLVs). The Text Chunks are linked to the CLVs.}
    \label{fig:kg_structure_n}
\end{figure}

This creates a clear separation of concerns. The CTV represents the conceptual \textbf{substance} of the component at a point in time, while the CLV represents \textbf{the specific wording} used to express it in a particular language. Consequently, the text chunk/segment itself—which we term a \textit{Text Unit}—is logically and exclusively associated with the most specific layer: the \textbf{Language Version} node. This design ensures that every piece of retrievable content is unambiguously tied to both a semantic state (the CTV) and a linguistic expression (the CLV), as depicted in Figure~\ref{fig:kg_structure_n}. This process results in a knowledge graph that provides a verifiable, point-in-time "ground truth" of the law.

\begin{figure}[htbp]
    \centering
    \includegraphics[width=0.7\textwidth]{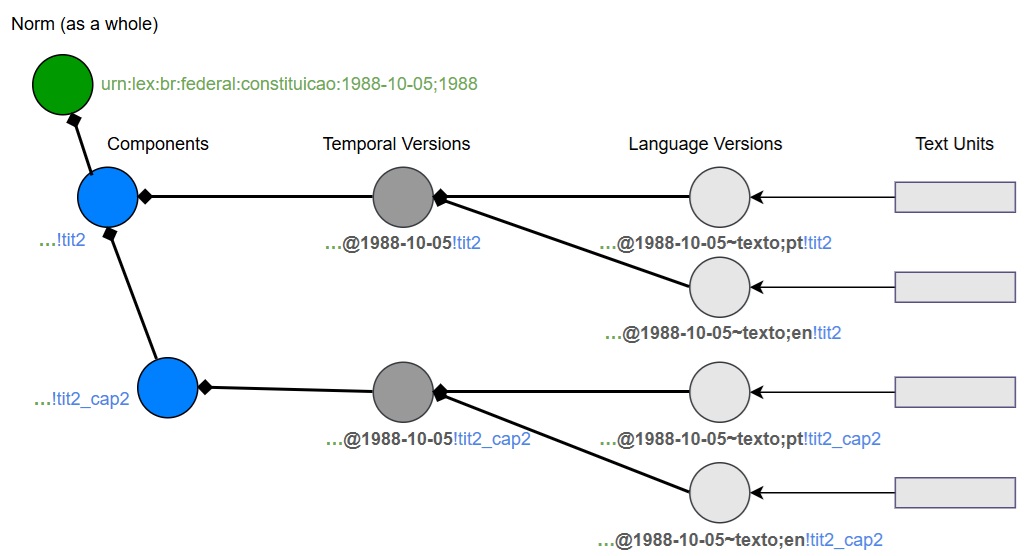} 
    \caption{Representation of the multilingual content (in Portuguese and in English).}
    \label{fig:kg_multilingual_n}
\end{figure}

The architecture’s efficiency becomes particularly evident when handling multilingual content. Assuming the initial Portuguese version has already been processed as described, let us consider what happens when an official English translation of the same temporal version is incorporated into the graph. The existing abstract and temporal structures—the \textbf{Norm (Work)}, \textbf{Component (Work)}, and date-stamped \textbf{Temporal Version (TV/CTV)} nodes—would remain entirely untouched. The process would only require the creation of a new set of language-specific nodes: a single new \textbf{Language Version (LV)} for the norm (in English), and corresponding \textbf{Component Language Version (CLV)} nodes for each component, all linked back to their pre-existing, language-agnostic CTVs. The new English \textit{Text Units} would then be associated exclusively with these new English CLVs, as depicted in Figure~\ref{fig:kg_multilingual_n}. This clearly demonstrates the model's elegant separation of a norm's conceptual and temporal identity from its linguistic expression, allowing the knowledge graph to scale efficiently to multiple languages without duplicating core structural information.

\subsection{Temporal Versions as Aggregations, Not Compositions}
\label{sec:aggregation}

Another innovation of our framework lies in the way it models the propagation of changes across the legal hierarchy. When a single component is amended on a given date, a new Temporal Version (CTV) is created for it. This local change necessitates a corresponding update to its ancestor components, propagating up to the \emph{Norm} itself, to reflect a new consolidated state on that date.

A naive approach would be to model this as a \textbf{Composition}, where a new parent CTV would be composed of newly created CTVs for all its children, even those whose text remained unchanged. This method is highly inefficient, creating vast amounts of redundant data and obscuring which components were actually modified.

\begin{figure}[htbp]
    \centering
    \includegraphics[width=0.7\textwidth]{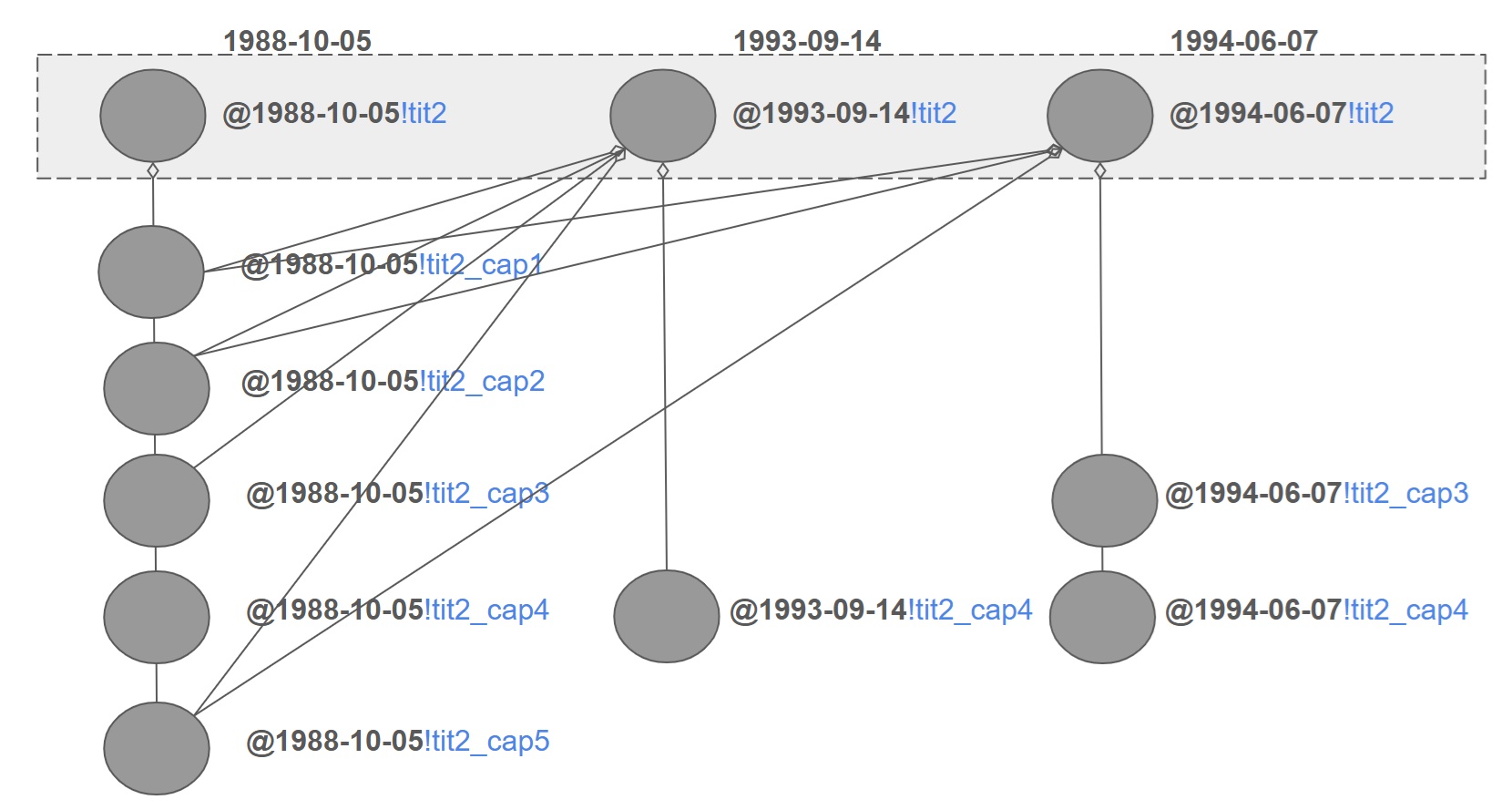}
    \caption{New \emph{Temporal Versions} of the component "tit2" (Title II) derived from new CTVs of some of its children (chapters). Unchanged child components have their most recent CTV reused (e.g., the 1988-10-05 CTV of tit2\_cap2 is aggregated into the 1993-09-14 CTV of tit2).}
    \label{fig:aggregator_versions}
\end{figure}

Instead, we model new parent CTVs as an \textbf{Aggregation}. A new parent CTV on date $D_n$ is formed by aggregating the \textit{most recent available} CTVs of each of its children. This means the new parent CTV points to the newly created CTV of the amended child and simply reuses the pre-existing, older CTVs of all unchanged children (Figure~\ref{fig:aggregator_versions}).

\begin{figure}[htbp]
    \centering
    \includegraphics[width=0.5\textwidth]{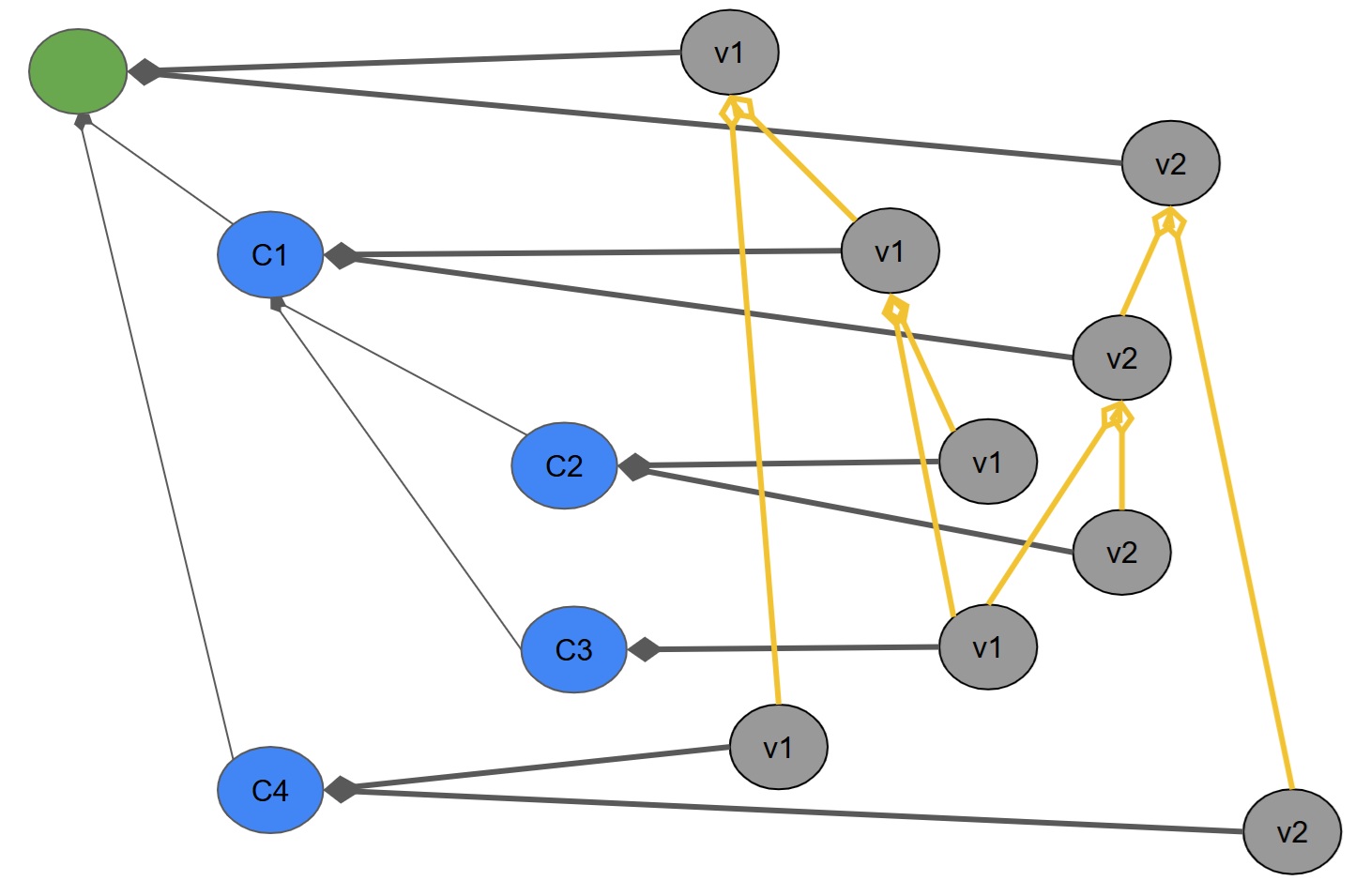}
    \caption{Diagram of aggregation relationships (orange) between \emph{Temporal Versions}, illustrating how child CTVs are reused by multiple parent CTVs at different times. For instance, the `v2` Temporal Version of `c1` aggregates the new `v2` CTV from Component `C2` but reuses the pre-existing `v1` CTV from the unchanged Component `C3`.}
\label{fig:aggregations_graph}
\end{figure}

This aggregation model provides an economical, non-ambiguous, and efficient representation of the norm's evolution. It establishes that a child's Temporal Version is not exclusively owned by a single parent version but can be reused across multiple parent versions at different points in time (Figure~\ref{fig:aggregations_graph}). 

In contexts like Retrieval-Augmented Generation (RAG), this mechanism is critical: it allows a system to deterministically reconstruct the full, correct text of any legal provision as it existed on a specific date by retrieving the appropriate aggregation of child versions, enabling complex and accurate temporal queries.

\subsection{Modeling Causality and Metadata as Retrievable Units}
\label{sec:actions_metadata}

To enrich the semantic capabilities of our graph, we extend the model beyond the textual content of norms. We introduce retrievable \textit{Text Units} for two additional types of information: normative events and structured metadata.

\subsubsection{Normative Events as Actions}
\label{sec:events_as_actions}

The lifecycle of a \emph{Temporal Version}—its creation and termination—is driven by a legislative event. To explicitly model this causality, we introduce an \textbf{Action} node for each granular change, grounded in event-centric legal ontologies \cite{demartim2025temporal, lima2008ontology}. An \textit{Action} node represents a specific command from a normative instrument, such as an amendment, a repeal, or an original enactment.\footnote{The structure of the causal link varies by the type of action. For an \textbf{amendment}, the \texttt{Action} connects the source provision (the instrument), the terminated CTV (the old version), and the created CTV (the new version). For an \textbf{original enactment}, however, there is no preceding version to terminate nor a separate instrument dictating the change; the \texttt{Action} simply represents the creation event that produces the initial CTV of a new legal norm, including its components.} It acts as a formal causal link, connecting key entities such as: 1) the source provision that dictates the change, 2) the \emph{Temporal Version} it terminates (if applicable), and 3) the new \emph{Temporal Version} it creates.

\begin{figure}[htbp]
    \centering
    \includegraphics[width=0.8\textwidth]{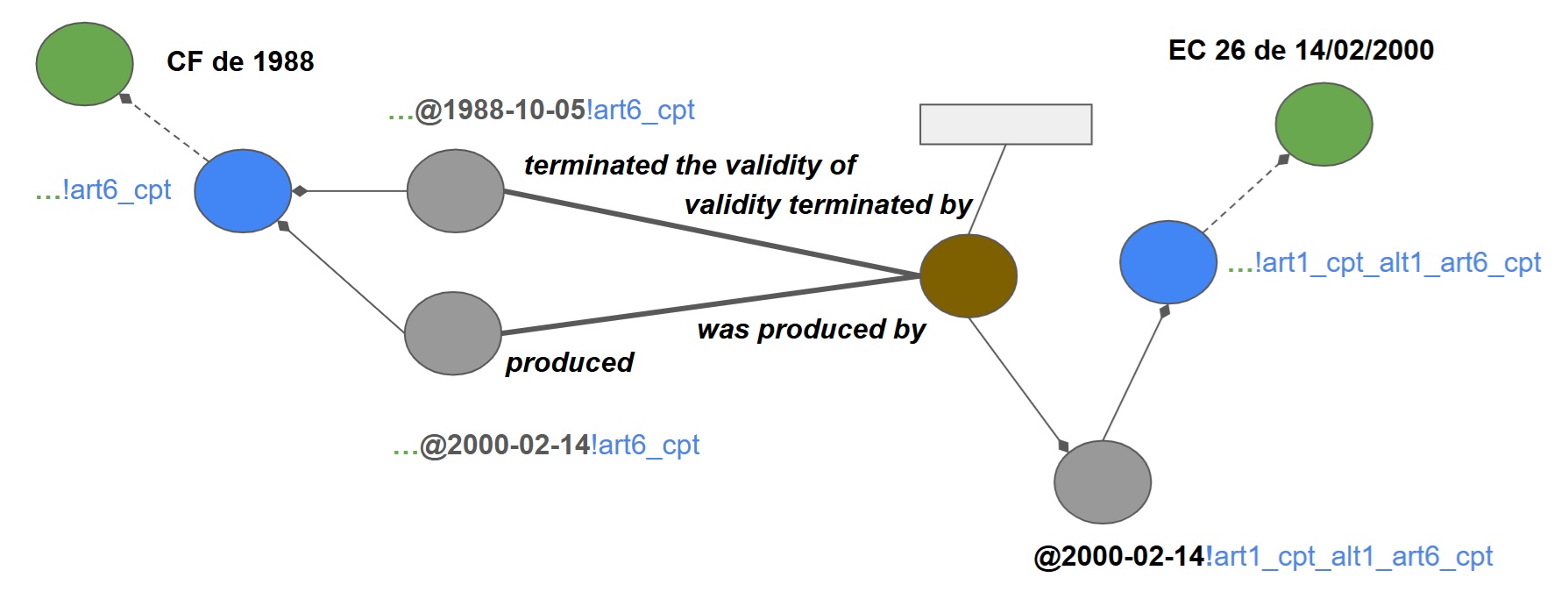}
    \caption{Representation of a legislative \emph{Action} (Event) in the knowledge graph. The \emph{Action}, commanded by the Art.1's caput of the Brazilian Constitutional Amendment 26, terminates the validity of the original 1988-10-5 CTV of Art. 6's caput of the Brazilian Federal Constitution of 1988 and produces its new 2000-02-14 CTV.}
    \label{fig:text_change_action}
\end{figure}

For each \textit{Action} node, we also generate a descriptive \textit{Text Unit}. This text is a structured, natural language summary of the event (e.g., \textit{Constitutional Amendment no. 26, of February 14, 2000, through the caput of its Art. 1º, provided a new wording for the caput of Art. 6º of the Brazilian Federal Constitution of 1988. This alteration terminated on 2000-02-14 the validity of the original version of this provision (from 1988-10-05) and established a new version effective from 2000-02-15, whose text became: 'Social rights include education, health, work, housing, leisure, security, social security, protection of motherhood and childhood, and assistance to the destitute, in the manner prescribed by this Constitution.'}). 

While the command for this change is embedded within the amending norm's text, creating this explicit, summary-level \textit{Text Unit} makes the legislative event itself a first-class, semantically searchable object. An embedding of this text allows an RAG system to directly retrieve the event, enabling it to answer queries not just about the content of the law, but also about its history, evolution, and the specific acts that caused changes.

\subsubsection{Metadata as Text Units for Multi-Aspect Retrieval}
A legal norm's identity is defined by more than just its textual content. Critical information—such as its publication date, alternative titles, or its relationships to other laws (e.g., succession or correlation)—is typically stored as structured metadata, making it invisible to standard text-based retrieval. To make this rich contextual information fully accessible to the RAG process, we convert it into natural language \textbf{Metadata Text Units}.

For each key entity in our graph—Norm, Component, or Temporal Version—we generate distinct \textit{Text Units} for each of its structured properties or informative relationships. An informative relationship, unlike a versioning \textit{Action}, describes a connection that does not create a new \textbf{version} of an existing text, but rather relates distinct entities. For instance, the succession relationship between two \textbf{Norms (Works)} would be reified into a dedicated Text Unit: \textit{"The 1967 Constitution of Brazil succeeded the 1946 Constitution of the United States of Brazil."}. This event terminates the validity of the former \textit{Work} as a whole, rather than creating a new version of it. Similarly, a simple metadata property would be textualized: \textit{"The 1988 Constitution of Brazil was published on October 5, 1988."}.

\begin{figure}[htbp]
    \centering
    \includegraphics[width=0.7\textwidth]{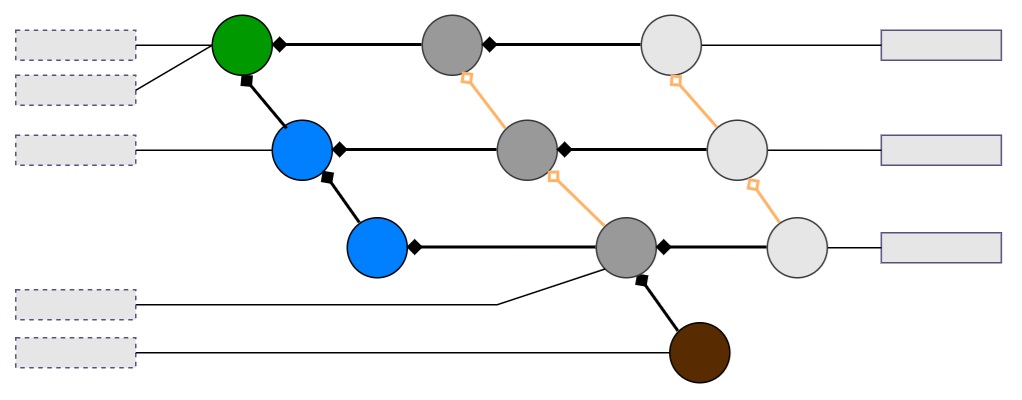} 
    \caption{Knowledge graph illustrating how \emph{Text Units} are derived from two sources: from \emph{Language Versions} (representing content) and from the other entities (\emph{Norm, Component, Temporal Version, Action}), representing metadata and relationships.}
    \label{fig:metadata_text_units}
\end{figure}

This approach operationalizes the concept of multi-aspect embeddings for the legal domain \cite{Park2020Unsupervised}. Instead of representing a legal entity with a single, monolithic vector, we generate multiple embedding vectors for it—one for its textual content, one for each causal \textit{Action}, and several for its various metadata properties and informative relationships. As illustrated in Figure~\ref{fig:metadata_text_units}, this multi-aspect representation allows a retrieval system to match queries against different facets of a norm's identity. A user can find a law based on its content, its properties, or its connections to other laws, providing multiple, complementary pathways to the most relevant information.

\subsection{Structure-Aware Retrieval via Curated Communities}
\label{sec:retrieval}

A key advantage of our graph-based model is its ability to enable structure-aware retrieval. While the original Graph RAG proposal relies on detecting communities algorithmically, our framework leverages two forms of curated, semantically meaningful communities that are intrinsic to the legal domain.

\begin{figure}[htbp]
    \centering
    \includegraphics[width=0.5\textwidth]{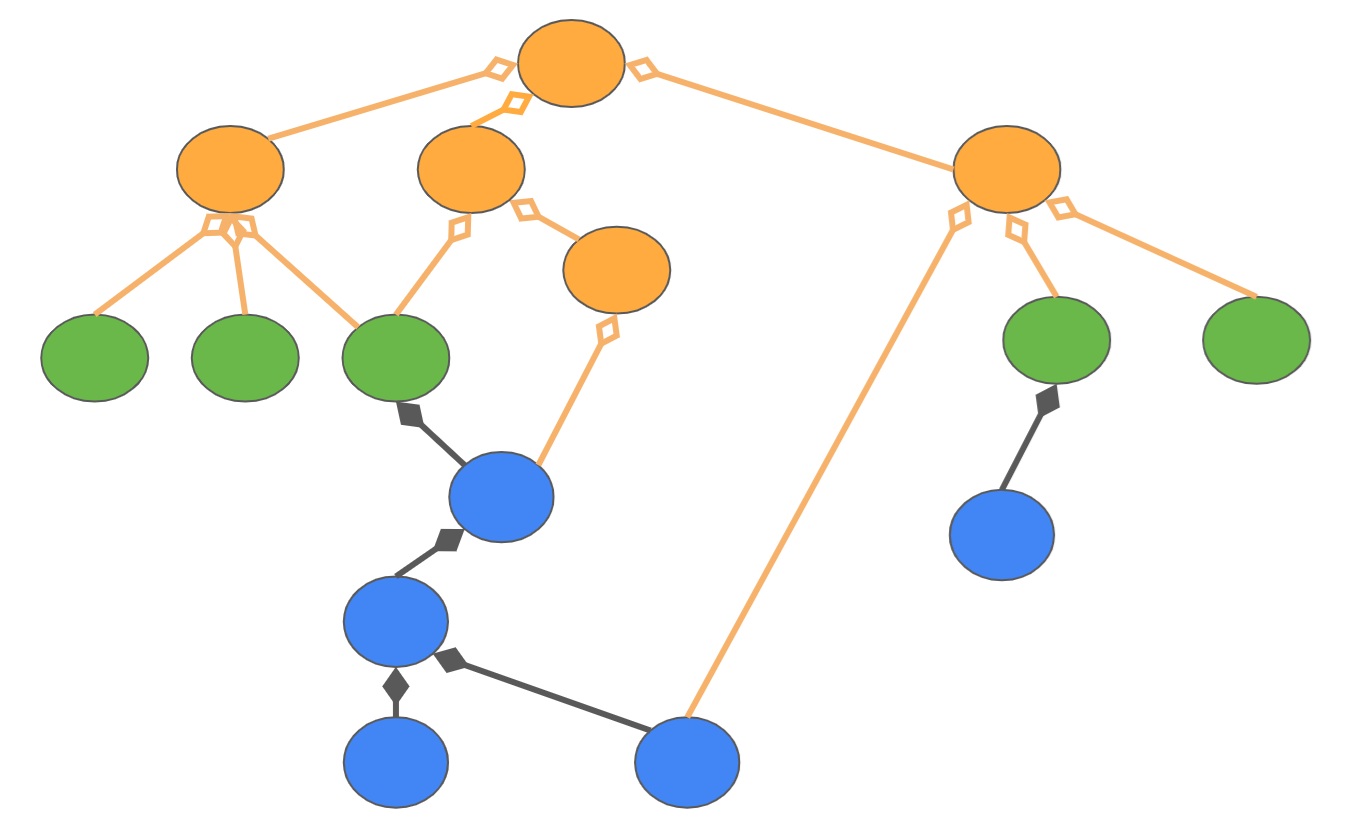}
    \caption{Diagram illustrating inter-norm (and component) aggregation by legal \emph{Theme} entities/nodes (orange), represented as higher-level communities that group \emph{Norm}, and eventually \emph{Component}, (Work) entities in the knowledge graph.}
    \label{fig:theme_aggregation}
\end{figure}

\begin{enumerate}
    \item \textbf{Internal Hierarchy (Structural Communities):} The predefined structure of a legal norm (Titles, Chapters, Sections) provides a natural, nested hierarchy. Each grouping component (e.g., a "Title") serves as a community that contains all its descendant components.
    
    \item \textbf{External Thematic Classification (Topical Communities):} Legal information systems often rely on human-curated taxonomies or thesauri to classify norms and provisions under established legal themes (e.g., "Social Security," "Environmental Law"). We model these external classifications by creating high-level \textbf{Theme} nodes. To make these themes themselves discoverable via semantic search, each \emph{Theme} node is associated with its own \emph{Text Unit}, containing a human-authored description of the topic (e.g., \textit{"Laws, articles, and provisions related to the protection of the environment, including regulations on pollution, conservation, and natural resources."}). These nodes are then linked to the relevant \textbf{Norm} and \textbf{Component} entities in our graph, forming cross-document, topically-coherent communities that can be queried directly (Figure~\ref{fig:theme_aggregation}).
\end{enumerate}

\begin{figure}[htbp]
    \centering
    \includegraphics[width=0.6\textwidth]{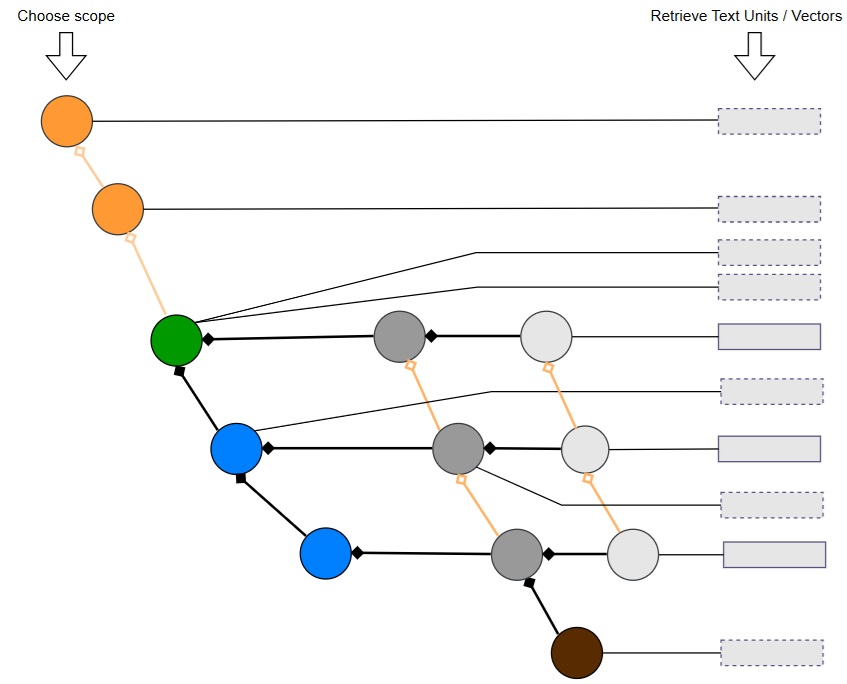}
    \caption{Diagram illustrating how a user can select a scope (e.g., a \emph{Theme}, \emph{Norm}, \emph{Component} or \emph{Version}) to filter the retrieval of relevant \emph{Text Units} from the knowledge graph.}
    \label{fig:retrieval_scope}
\end{figure}

This curated community structure enables a powerful filtering mechanism for retrieval. A user can define the scope of a query by selecting a specific entry point in the graph—be it a \textit{Theme}, a \textit{Norm}, a \textit{Component}, or even a specific \textit{Temporal Version}. The system first traverses the graph from this entry point to gather all associated Text Units (from the node itself and all its descendants). The semantic vector search is then performed exclusively on this pre-filtered, contextually relevant subset of Text Units (Figure~\ref{fig:retrieval_scope}).

This structure-aware filtering transforms retrieval from a flat search across an entire corpus into a process of semantic navigation. This approach is not only more computationally efficient but also yields far more precise results, as the search is constrained a priori to a contextually relevant subgraph defined by the logical and thematic structure of the law itself.

\section{Case Study: Modeling and Querying Legal Evolution}
\label{sec:case_study}

To demonstrate the practical viability and advantages of our framework, we present a case study centered on the evolution of the Brazilian Federal Constitution of 1988. With over one hundred amendments since its enactment, its complex diachronic history serves as an ideal stress test for any temporal modeling approach.

This section illustrates how our graph-based model, when serving as the knowledge backbone for a Retrieval-Augmented Generation (RAG) system, enables three critical capabilities. First, it facilitates deterministic point-in-time retrieval, pinpointing the exact version of a provision valid on any date. Second, it supports hierarchical impact analysis, aggregating changes across entire structural sections of a legal norm. Third, it enables auditable provenance reconstruction, tracing the complete causal lineage of any textual element. We will show how these capabilities provide a level of reliability and analytical depth that is inherently unattainable for standard RAG systems operating on flat, unstructured legal corpora.

\subsection{Dataset and Conceptual Architecture}
\label{sec:implementation}
The case study uses the official texts of the Brazilian Federal Constitution (1988) and all subsequent Constitutional Amendments enacted as of early 2024. The documents would serve as input for a pipeline that preprocesses them into structure-aware segments (titles, articles, paragraphs, items). A potential implementation of our framework would integrate three core components:
\begin{enumerate}
    \item \textbf{A Relational-Vector Knowledge Store:} For this study, our knowledge base could be implemented in a modern relational-vector database such as \textbf{Oracle Database 23ai}. The graph structure of our ontological model could be realized over the relational model: \textit{Theme}, \textit{Norm\_Component}, \textit{Temporal Version}, \textit{Language Version}, and \textit{Action} entities would be stored as tables, and their relationships (e.g., hierarchy, versioning) represented by auxiliary tables and foreign key constraints. The \textit{Text Unit} table would include a dedicated vector column to store the text embedding alongside its content. This architecture demonstrates that the proposed graph model can be effectively implemented on such databases, without requiring a dedicated property graph engine.
    \item \textbf{An Embedding Model:} The vector representations for all \textit{Text Units} could be generated using a powerful open-source model such as \textbf{Qwen3 Embedding}. Such a model is a strong candidate not only for its performance on multilingual text comprehension but also because it can be self-hosted, ensuring data privacy and control.`

    \item \textbf{An LLM-based Generation Module:} The final synthesis of natural language responses, based on the retrieved context, could be performed by a model like Google's Gemini 1.5 Flash, selected for its long-context capabilities and efficiency.
\end{enumerate}

\noindent While this conceptual architecture could leverage the efficiency of a converged database, the framework itself is tool-agnostic. It could be readily implemented with other combinations of dedicated graph databases (e.g., Neo4j), vector stores (e.g., Pinecone, FAISS), and LLMs.

The knowledge graph would be populated through a semi-automated pipeline. This process would involve processing the official, multi-temporal compiled version of the Constitution (and subsequent Constitutional Amendments), applying the semantic segmentation rules outlined in Section~\ref{sec:ontology_construction}. The legislative events described in each Constitutional Amendment would then be modeled as \textit{Action} nodes, linking the source provisions to the target components they modified. This process would result in a fine-grained, diachronic representation of the entire legislative history of the Constitution.

\subsection{Modeling a Legislative Amendment: The Case of Article 6}
\label{sec:modeling_art6}
\begin{figure}[htbp]
\centering
\includegraphics[width=0.9\textwidth]{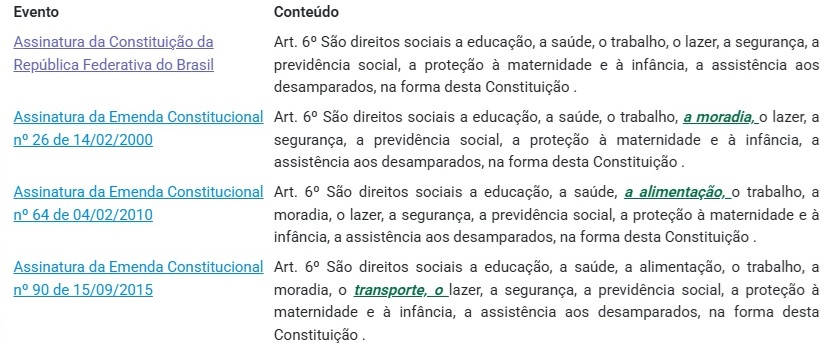}
\caption{Original Version and subsequent Versions of Article 6 of Brazilian Constitution generated by 3 Constitutional Amendments}
\label{fig:art6_versions}
\end{figure}

To illustrate the application of our model, we focus on the caput of Article 6 of the Constitution, which enumerates social rights. This provision is an ideal test case, having been amended multiple times, as showed in Figure \ref{fig:art6_versions}. 

We will analyze its state immediately before and after the enactment of Constitutional Amendment (CA) No. 26 of February 14, 2000, which added the right to "housing" (\textit{a moradia}) to the list. Following our framework, this legislative event is modeled as follows:
\begin{itemize}
\item An \textbf{Action} node is created to represent the amendment command from CA No. 26. A descriptive \textit{Text Unit} is generated for this action, summarizing its effect, as exemplified in Section \ref{sec:events_as_actions}.
\item This \textit{Action} formally \textbf{terminates the validity} of the original \emph{Temporal Version} (CTV) of Art. 6's caput, dated 1988-10-05.
\item The \textit{Action} also \textbf{produces} a new \emph{Temporal Version} (CTV) for Art. 6's caput, dated 2000-02-14.

\item A new \emph{Language Version} (CLV) is created as a linguistic realization of this new CTV. This CLV node is then associated with the definitive \textit{Text Unit} containing the updated wording of the article's caput.

\item Following the aggregation model (Section~\ref{sec:aggregation}), new Temporal Versions are propagated up the hierarchy (for its Chapter, Title, etc.), reusing the unchanged versions of sibling components.

\end{itemize}

As a result, the graph would contain a deterministic and fully traceable record of this specific legislative change. The cause (the command from CA No. 26, represented by the Action node) is explicitly linked to its precise effect: the termination of one textual state and the creation of another. This fine-grained, event-centric representation, previously illustrated in Figure~\ref{fig:text_change_action}, is the key to enabling the accurate temporal queries we demonstrate next.

\subsection{Qualitative Evaluation: Answering Complex Temporal Queries}
\label{sec:evaluation}
The ultimate value of our framework lies in its ability to answer complex legal queries that are contingent on the law's precise state at a specific point in time. We evaluate this capability qualitatively by contrasting the designed behavior of our ontology-driven system with that of a "standard RAG" baseline. For this case study, a standard RAG is defined as a system operating on a flat index of text chunks derived from the \textit{current} version of the Constitution, lacking any explicit model of the document's diachronic structure.

\subsubsection*{Query Pattern 1: Point-in-Time Retrieval (Generic Strategy)}
\textbf{User Query (example):} \textit{"What were the social rights listed in Article 6 of the Brazilian Constitution in 1999?"}

\paragraph{Baseline RAG (temporally-naïve)}
Without temporal awareness, a standard RAG system would retrieve the \emph{current} text most semantically similar to the query. Even when the vector index also contains the contemporary texts of amending acts that describe changes to Article~6, retrieval remains unconstrained by legal validity intervals; the LLM therefore would tend to conflate contemporary descriptions with the target historical date. In practice, it would surface modern content and produce anachronistic results (e.g., including rights such as housing ("a moradia") or food ("a alimentação") that were introduced later. This illustrates a known limitation of flat retrieval systems that lack an explicit, versioned representation of diachronic change.

\paragraph{Ontology-Driven SAT Graph RAG (generic, deterministic)}
Our ontology-driven SAT Graph RAG framework is designed to resolve this class of queries through a deterministic multi-step plan, which generalizes to any legal provision and any temporal scope:

\begin{enumerate}
\item \textbf{Query planning and structured entity extraction.}
A schema-aware LLM acts as a query planner, converting the natural language request into structured constraints:
\begin{itemize}
    \item \textbf{Structural target:} maps ``Article 6 of the Brazilian Constitution'' to its canonical \texttt{ComponentID}, disambiguated by jurisdiction and source norm (e.g., \url{urn:lex:br:federal:constituicao:1988-10-05;1988!art6}). Alias handling and normalization ensure robustness across different surface forms.
    \item \textbf{Temporal scope:} interprets ``in 1999'' as a validity constraint for the date $t$ (e.g., any $t \in [1999\text{-}01\text{-}01,1999\text{-}12\text{-}31]$). This constraint is formalized as an interval condition, avoiding reliance on arbitrary cutoffs such as 31 December. If the query lacks an explicit temporal constraint (e.g., asking "what \textit{are} the rights..."), the planner defaults the temporal scope to the current system date (t = now()), effectively retrieving the most up-to-date version.
    \item \textbf{Language and format:} optional constraints specify preferred language or granularity (e.g., caput vs. full list of items).
\end{itemize}

\item \textbf{Temporal scoping via graph traversal.}
 This is the core deterministic step. Starting from the identified \texttt{ComponentID} ("Art. 6"), the system first resolves the full hierarchical scope by traversing downwards to gather the entry component itself and all its text-bearing descendants (e.g., the nodes for `Art. 6`, its `caput`, and any `items`). With this set of target components defined, the system traverses their chain of Component Temporal Versions (CTVs) to select the versions whose validity interval satisfies:
\[
tv.valid\_start \leq t < coalesce(tv.valid\_end, +\infty).
\]
This step yields the precise set of CTVs that represent the complete state of Article 6 and its parts in 1999. To ensure determinism when multiple CTVs exist within the year, the system applies a temporal-resolution policy (default \(\operatorname{SnapshotLast}\)): it chooses \(t^\star=\sup\{t \in [t_1,t_2]\}\) and returns the CTV valid at \(t^\star\). The policy is disclosed with the answer.
This step is both efficient and unambiguous thanks to the aggregation model (Section~\ref{sec:aggregation}), where temporal versions of unchanged child components are reused rather than duplicated, ensuring a single, canonical representation for any component at any point in time.

\item \textbf{Retrieval of the historically valid text units.}
From the set of valid CTVs identified in the previous step, the system retrieves the corresponding Component Language Versions (CLVs) in the requested language, and selects the associated \textit{Text Units} that are relevant to the query scope.

\item \textbf{Fact-grounded generation.}
The temporally valid Text Units are then passed to the generation module. For enumerative queries, the LLM is instructed to produce copy-based (extractive) answers that preserve the original enumeration and attach explicit citations to the CTV/CLV. Optionally, the system may also expose the \textbf{Action} node that initiated or terminated the version, supporting explainability.
\end{enumerate}

\noindent\textbf{Exemplary Answer (for 1999).}
\textit{``In 1999, the social rights listed in Article 6 of the Brazilian Constitution were education, health, work, leisure, security, social security, protection of motherhood and childhood, and assistance to the destitute.''}
[cited from the CTV valid in 1999].

\paragraph{Robustness notes (scalable beyond the example).}
This strategy naturally supports: (i) temporal intervals ($t \in [t_1,t_2]$), (ii) distinctions between promulgation and validity dates, (iii) multilingual retrieval with fallback policies, and (iv) querying diverse scopes simply by changing the entry node. This includes scopes defined by the document's internal hierarchy (e.g., from a specific item to entire Chapters or Titles) as well as by external, cross-document thematic classifications (e.g., starting from a 'Social Security' Theme node). The algorithm remains unchanged, handling even queries about the current state of the law by defaulting the temporal scope to the present date, thus ensuring scalability.

\subsubsection*{Query Pattern 2: Hierarchical Impact Analysis}
\textbf{User Query (example):} \textit{"Provide a summary of all textual changes made to the components within Chapter II ('On Social Rights') of the Brazilian Constitution after the year 2010."}
\paragraph{Baseline RAG (structurally-naïve)} A conventional flat RAG system is fundamentally incapable of answering this query reliably. It lacks the explicit representation of the document's hierarchy needed to even identify which provisions belong to "Chapter II". Any attempt to answer would rely on brittle, heuristic searches and complex reasoning, making a complete and factually correct summary highly improbable.

\paragraph{Ontology-Driven SAT Graph RAG (structure-aware)}
Our framework is designed to handle such structural-temporal queries by treating the legal hierarchy as a primary queryable structure. The process is deterministic:

\begin{enumerate}
    \item \textbf{Scope Identification and Policy Selection:} The query planner first identifies the structural target (the \texttt{Component} node for "Chapter II") and the temporal window ("after 2010"). It also selects a \textbf{scope-membership policy} to handle potential structural reorganizations over time. The default is a \textbf{"snapshot-anchored"} approach, which defines the scope as all components that belonged to Chapter II at the beginning of the time window. Alternative policies (e.g., action-time, lifetime) can be used for different analytical needs, and the chosen policy is always disclosed with the result.

    \item \textbf{Descendant Component Retrieval:} With the entry node and membership policy defined, the system executes the retrieval. It traverses the graph downwards from the "Chapter II" node, gathering the set of all descendant components (e.g., Art. 6, Art. 7) that satisfy the membership policy selected in the previous step.

    \item \textbf{Causal Event Aggregation:} For this set of components, the system gathers all legislative events that drove their evolution within the specified time window. This is achieved by retrieving all \texttt{Action} nodes that acted upon the Component Temporal Versions (CTVs) associated with these components.

  \item \textbf{Hierarchical Summary Generation:} The descriptive \textit{Text Units} of the retrieved \texttt{Actions} are aggregated. The LLM then receives this structured context with instructions to synthesize a hierarchical summary, grouping changes by the component they affected and providing a top-level timeline of all impact dates.
\end{enumerate}

\paragraph{Exemplary Output (illustrative)}
\begin{quote}\small
\textbf{Impact Summary for Chapter II (2010-2019):}
\begin{verbatim}
+-- Art. 6 (caput): 2 amendments
|   +-- CA 64/2010: added "food"
|   '-- CA 90/2015: added "transportation"
'-- Art. 7: 1 amendment
    '-- CA 72/2013: extended domestic workers' rights
\end{verbatim}
\textbf{Chapter-level impact dates:} \{2010-02-04, 2013-04-02, 2015-09-15\}.
\end{quote}
This demonstrates the ability to move beyond simple fact retrieval to perform complex impact analysis, a capability directly enabled by the explicit modeling of the law's hierarchical and diachronic structure.

\subsubsection*{Query Pattern 3: Provenance and Causal-Lineage Reconstruction}
\textbf{User Query (example):} \textit{"Trace the full legislative lineage that introduced the term 'food' in Article 6 of the Brazilian Constitution."}`
\paragraph{Baseline RAG (provenance-naïve)}
A conventional flat RAG system is poorly suited to this task. It may find text fragments mentioning the phrase, but it cannot reliably attribute the change to a specific legal instrument or reconstruct the chronological sequence of edits. Provenance requires linking concrete text spans across successive versions and associating each edit with a discrete \textbf{Action}; heuristic retrieval alone is insufficient.

\paragraph{Ontology-Driven SAT Graph RAG (strategy-aware)}
Our framework's query planner would select the most efficient execution strategy based on the query's constraints. As this query provides both a structural ("Article 6") and a textual ("food") target, the planner would opt for a \textbf{structure-first} approach to narrow the search space:

\begin{enumerate}
    \item \textbf{Query Analysis and Constraint Extraction:} The planner first parses the user's query into a set of structured constraints, without committing to an execution order. This includes identifying and canonicalizing the \textbf{structural target} (`ComponentID` for "Art. 6") and the \textbf{textual target} ("food").

    \item \textbf{Hierarchical Scope Resolution:} The system takes the identified structural target (`ComponentID` for "Art. 6") and resolves the full search scope. This scope includes the component itself and the entire descendant subtree beneath it in the legal hierarchy. This ensures that if a user specifies a high-level component (e.g., "Article 6"), the search will correctly include all its children (caput, paragraphs, items, etc.).

    \item \textbf{Constrained Span Location:} Now, instead of searching the entire corpus, the system performs a targeted search for the textual target ("food") only within the version history of the components in the resolved hierarchical scope. Using lexical and semantic indexes, it locates all relevant \textit{Text Units}.

    \item \textbf{Causal Action Identification:} For each located text span, the system traverses the graph to find the \textbf{Action} node that created the \textit{Temporal Version} (CTV) in which the text appears.

    \item \textbf{Causal Chain Assembly:} The system then traces the lineage of the specific component where the text was found by following the chain of \texttt{Action} nodes backward in time, assembling a deterministic and ordered Directed Acyclic Graph (DAG).

    \item \textbf{Provenance Report Generation:} The ordered chain of \texttt{Actions} and their associated "before" and "after" text snippets are passed to the LLM to synthesize a chronological narrative with auditable citations and a machine-readable annex.
\end{enumerate}

\paragraph{Note on Planner Flexibility}
This illustrates the planner's adaptability. Had the user asked a broader question without a structural scope (e.g., \textit{"Trace the lineage of the term 'food' in social rights"}), the planner, seeing no structural constraint, would have automatically switched to a \textbf{span-first} strategy. It would first locate all occurrences of "food" across the entire corpus (Step 3, but unconstrained) and then use the graph to reconstruct the provenance for each distinct location found (Steps 4-6).

\paragraph{Exemplary Output (Abridged)}
\begin{quote}
\small
\textbf{Provenance Report: "food" in Art. 6 caput}
\begin{itemize}[leftmargin=*, noitemsep, topsep=0.5em]
    \item \textbf{Pre-State:} Valid until 2010-02-03 (no mention of "food").
        \begin{itemize}[noitemsep]
            \item \textit{Last change by:} CA 26/2000. \textit{Source CTV:} \texttt{...2000-02-14!art6\_cpt}.
        \end{itemize}
    \item \textbf{Causal Event:} Action from \textbf{CA 64/2010} (effective 2010-02-04).
        \begin{itemize}[noitemsep]
            \item \textit{Effect:} Inserted the term "food".
        \end{itemize}
    \item \textbf{Post-State:} Valid from 2010-02-04.
        \begin{itemize}[noitemsep]
            \item \textit{Source CTV:} \texttt{...2010-02-04!art6\_cpt}.
        \end{itemize}
    \item \textbf{Audit Trail:}
        \begin{itemize}[noitemsep]
            \item \textit{Causal Chain:} [\textbf{Action}(CA 26/2000)] \(\rightarrow\) [\textbf{Action}(CA 64/2010)].
            \item \textit{Match Confidence:} Exact (1.0). \textit{Annex:} Full JSON record available.
        \end{itemize}
\end{itemize}
\end{quote}

\noindent By making causality and versioning explicit graph structures, and by selecting the optimal query strategy, this pattern enables deterministic and auditable provenance reports that a baseline RAG cannot reliably produce.

\subsection{A unified execution strategy}
\label{sec:unified_strategy}
To support the three query patterns exemplified above (point-in-time retrieval; hierarchical impact analysis; provenance reconstruction) we propose a single, planner-guided execution strategy that is both deterministic and configurable. The strategy centralizes query interpretation and then applies a small set of composable steps:
\begin{enumerate}
    \item \emph{canonicalization} of structural, temporal and textual constraints; 
    \item \emph{scope resolution} over the legal hierarchy (with a declared membership policy); 
    \item \emph{strategy selection} (structure-first / span-first / time-first) by the planner; 
    \item \emph{deterministic CTV selection} according to a stated temporal policy (e.g., SnapshotLast); 
    \item \emph{scoped retrieval} of Text Units (structural + temporal filters followed by vector/lexical ranking); 
    \item \emph{causal aggregation} of Action nodes, and their associated descriptive Text Units, when required; \item \emph{provenance chain assembly} (DAG of Actions) and 
    \item \emph{fact-grounded generation} with explicit disclosure of the policies and a machine-readable annex (JSON) containing the provenance trace and confidence scores.
\end{enumerate}

This unified pipeline is deliberately modular. For point-in-time queries the planner typically executes steps 1, 2, 3, 4, 5, 8 (canonicalize → resolve scope → select strategy → select deterministic CTV → retrieve Text Units → fact-grounded generation). For hierarchical impact analysis the planner usually follows 1, 2, 3, 6, 5, 8 (canonicalize → resolve scope → select strategy → aggregate Actions across descendants → retrieve representative Text Units/snapshots if required → generate hierarchical summary). For provenance reconstruction the planner executes 1, 2, 3, 5, 6, 7, 8 (canonicalize → resolve scope → select strategy → locate spans/TextUnits across relevant CTVs → identify Actions affecting those spans → assemble ordered DAG of Actions → generate provenance report). The planner records the chosen policies (membership, temporal policy, retrieval $k$, etc.) with every response to ensure auditability and reproducibility.

\paragraph{Operational Defaults and Disclosures.} A reference implementation of this framework should define and report its operational defaults to ensure reproducibility. For instance, it could specify parameters such as the embedding model and dimension (e.g., Qwen-3 with 256 dimensions), the similarity function and retrieval depth (e.g., cosine similarity with $k=8$), and the default temporal policy (e.g., \(\operatorname{SnapshotLast}\)). Fundamentally, all answers generated by the system should explicitly state the temporal and membership policies used for that specific query and attach a machine-readable JSON provenance annex when \texttt{Action} nodes are part of the result. These defaults should be configurable, and their disclosure is essential to preserve determinism and user trust.

\paragraph{Benefits.} By centralizing planning and exposing policy choices, the proposed design makes it possible to achieve (i) deterministic point-in-time answers, (ii) scalable hierarchical aggregations, and (iii) auditable provenance reconstruction—all from the same core pipeline. This design keeps the system predictable for legal usages while remaining extensible to other jurisdictions and corpus types.

\section{Discussion}
\label{sec:discussion}

The case study's query patterns demonstrate that our ontology-driven framework provides a substantially more capable and deterministic retrieval substrate than flat, temporally-naïve RAG systems. The unified, planner-guided execution strategy shows how a single set of composable operations can resolve complex queries for point-in-time retrieval, hierarchical impact analysis, and auditable provenance. By making policies for temporal and structural resolution explicit (e.g., \(\operatorname{SnapshotLast}\), snapshot-anchored membership), the framework ensures that its outputs are reproducible, explainable, and traceable back to specific graph entities—a core contribution to the field of explainable legal AI~\cite{richmond2024explainable}.

\paragraph{Scalability and Maintenance}
Applying this framework at a national scale requires significant engineering and governance. Key strategies include combining automated ingestion pipelines with human-in-the-loop validation, pre-computing indexes and materializing common temporal snapshots to accelerate queries, and processing new legislation incrementally. These represent a strategic trade-off, investing in data curation and storage to gain the determinism and query performance essential for high-stakes legal applications.

\paragraph{Dependence on Upstream Data Quality}
The framework's determinism amplifies the advantages of structured data but also propagates upstream errors more visibly. Erroneous validity intervals or incorrect Action metadata will lead to systematically flawed retrievals. Mitigation is therefore essential and relies on practices such as multi-source corroboration, attaching confidence scores to all outputs, and implementing robust validation and rollback workflows.

\paragraph{Scope, Adaptability, and Generalization}
The proposed ontology is optimally suited to statutory corpora where hierarchy and amendment actions are explicit and canonical. Adapting this model to other domains, whether legal or non-legal, requires careful ontological and procedural adjustments, primarily centered on how "structure" is defined and "change" is tracked.

For \textbf{other legal domains}:
\begin{itemize}[noitemsep, topsep=0pt]
    \item \textbf{Case Law:} The focus would shift from textual diffs to modeling citation graphs, judicial hierarchies, and the temporal evolution of legal precedents (doctrinal drift).
    \item \textbf{Contracts/Regulations (within law):} Their structural heterogeneity may demand bespoke semantic segmentation rules for graph construction, as their internal organization can be less standardized than statutory law.
\end{itemize}

The framework's core principles—separating abstract structure from dated textual expressions and modeling change via explicit events—are, however, highly generalizable beyond the legal sphere. For \textbf{non-legal domains}:
\begin{itemize}[noitemsep, topsep=0pt]
    \item \textbf{Commercial Contracts:} These documents often evolve through amendments and restatements, making our versioning model highly applicable.
    \item \textbf{Technical Documentation:} Tracking software releases, API changes, and documentation versions (e.g., user manuals, specifications) also aligns well with our event-centric, policy-driven approach.
\end{itemize}

\paragraph{Evaluation and Metrics}
A rigorous quantitative validation of our framework's claims of determinism and accuracy requires a dedicated benchmark, which is currently a notable gap in the legal AI community. To address this, we propose the creation of a ground-truth testbed built upon a legislative corpus with a known, complex amendment history (such as the one used in our case study). This testbed would consist of a curated set of query-answer pairs designed to evaluate the core capabilities of any structure- and temporally-aware retrieval system. To support these claims, future work should focus on the following quantitative metrics:

\begin{itemize}
    \item \textbf{For Deterministic Temporal Retrieval (Query Pattern 1):}
    \begin{itemize}
        \item \textbf{Temporal Precision:} The fraction of retrieved Component Temporal Versions (CTVs) that are correctly valid for the target date or interval specified in the query.
        \item \textbf{Temporal Recall:} The fraction of all ground-truth relevant CTVs for a given query that the system successfully retrieved.
    \end{itemize}

    \item \textbf{For Hierarchical Impact Analysis (Query Pattern 2):}
    \begin{itemize}
        \item \textbf{Action-Attribution Accuracy:} The F1-score for correctly identifying all legislative \texttt{Action} nodes that acted upon the temporal versions (CTVs) of components within the specified hierarchical scope and time window.
        \item \textbf{Summary Completeness:} The percentage of ground-truth changes within the scope that are correctly included in the final generated summary.
    \end{itemize}

    \item \textbf{For Provenance Reconstruction (Query Pattern 3):}
    \begin{itemize}
        \item \textbf{Causal-Chain Completeness:} The percentage of the ground-truth causal chain (the sequence of \texttt{Action} nodes) that was fully and correctly reconstructed by the system.
    \end{itemize}
    
    \item \textbf{User-Centred Measures:} Beyond automated metrics, user studies are essential to measure the practical impact, such as the reduction in time-to-answer for complex legal research tasks and the perceived increase in the trustworthiness and auditability of the generated answers.
\end{itemize}

\noindent We advocate for the development and public release of such an annotated benchmark to enable reproducible evaluation and a fair comparison between different temporally-aware legal retrieval systems.

\paragraph{Implications for Legal Practice and Ethical Considerations}
For practitioners, this approach unlocks faster, auditable historical research and deterministic summaries of legislative change. However, we advise conservative operational policies. The framework is designed to produce \textbf{verifiable assistant outputs}, not definitive legal advice. Its primary role is to accelerate human expertise by providing candidate answers with full provenance exposed for independent verification. This responsible deployment model is guided by several key ethical considerations:

\begin{itemize}[noitemsep, topsep=0pt, leftmargin=*]
    \item \textbf{Policy Transparency:} The system must be transparent about its internal resolution policies (e.g., temporal and scope-membership rules). Users must understand \textit{how} an answer was constructed to properly assess its context and reliability.
    
    \item \textbf{Data Equity and Access:} The framework's effectiveness is contingent on the availability of high-quality, machine-readable legal data. A significant ethical risk is creating a "justice gap," where only well-funded jurisdictions can build such systems, potentially marginalizing those with less accessible legal corpora.
    
    \item \textbf{Curation and Auditability:} Maintaining immutable audit logs for all data curation and graph updates is essential. The process of modeling legal history must itself be transparent and traceable to prevent manipulation or the introduction of systemic errors.
\end{itemize}

\section{Conclusion}
\label{sec:conclusion}

This paper introduced the \textbf{Structure-Aware Temporal Graph RAG (SAT-Graph RAG)}, an ontology-driven architecture designed to overcome the critical limitations of standard retrieval systems when applied to diachronic legal corpora. Grounded in a formal, LRMoo-inspired model, our framework treats a document's formal structure, its temporal versions, and the legislative \texttt{Actions} that drive its evolution as first-class, interconnected entities. This enables a unified, planner-guided query strategy that transforms legal information retrieval from a probabilistic search into a deterministic, auditable process.

Our principal contributions are:
\begin{itemize}
    \item The \textbf{application of a formal, multi-layered representation}, based on~\cite{demartim2025temporal}, that separates abstract legal Works from their concrete Temporal and Language Expressions, enabling precise, point-in-time reconstruction.
    \item \textbf{An efficient version-aggregation strategy} that models new temporal states as aggregations of reusable components, preserving hierarchical integrity while avoiding data redundancy.
    \item The \textbf{reification of legislative \texttt{Actions} as first-class, retrievable entities}, which makes the causes of textual change directly queryable, attributable, and assemblable into auditable provenance chains, building upon the event-centric model from~\cite{demartim2025temporal}.
    \item \textbf{A multi-aspect retrieval strategy} that textualizes structured metadata and causal events into distinct \textit{Text Units}, making a norm's context—not just its content—a first-class, searchable facet.
    \item \textbf{A unified, policy-driven query framework} that leverages the graph's explicit formal structure to deterministically resolve complex requests for point-in-time retrieval, hierarchical impact analysis, and causal-lineage reconstruction.
\end{itemize}

While the approach requires a principled investment in data curation and governance, it lays a necessary foundation for the next generation of trustworthy legal AI. By making awareness of formal document structure a core principle alongside explicit models of temporality and causality, our framework offers a practical and robust path toward building AI systems that can reliably support high-stakes legal decision-making.

\subsection*{Future Work}
We identify several promising research directions, including the publication of a formal OWL ontology for the model and the development of annotated benchmarks for reproducible evaluation. Future work will also focus on extending the framework by adapting it to other domains and investigating hybrid models that combine our top-down, structural approach with bottom-up, content-driven entity graphs, as described in the original Graph RAG proposal~\cite{edge2024graphrag}. Finally, we envision the creation of interactive tooling for provenance visualization and legal workflow integration.

\bibliographystyle{ios1} 
\bibliography{referencias}

\end{document}